\documentclass[a4paper]{llncs}
\usepackage{amsmath}
\usepackage{amssymb}
\usepackage{cite}
\usepackage{graphicx}
\usepackage{grffile}
\usepackage[pdftex]{hyperref}
\usepackage{subfig}
\usepackage{tabularx}
\usepackage{booktabs}
\graphicspath{{figures/}}
\begin{document}

\mainmatter

\title{Recurrent Neural Networks for Aortic Image Sequence Segmentation with Sparse Annotations}
\titlerunning{Recurrent Neural Networks for Aortic Image Sequence Segmentation}

\author{Wenjia Bai\inst{1} \and Hideaki Suzuki \inst{2} \and Chen Qin \inst{1} \and
Giacomo Tarroni \inst{1} \and \\
Ozan Oktay\inst{1} \and Paul M. Matthews \inst{2,3} \and Daniel Rueckert \inst{1}
\\
\authorrunning{W. Bai et al.}}

\institute{Biomedical Image Analysis Group, Department of Computing,
  Imperial College London, UK
  \and
  Division of Brain Sciences, Department of Medicine,
  Imperial College London, UK
  \and
  UK Dementia Research Institute,
  Imperial College London, UK
}

\maketitle

\begin{abstract}
Segmentation of image sequences is an important task in medical image analysis, which enables clinicians to assess the anatomy and function of moving organs. However, direct application of a segmentation algorithm to each time frame of a sequence may ignore the temporal continuity inherent in the sequence. In this work, we propose an image sequence segmentation algorithm by combining a fully convolutional network with a recurrent neural network, which incorporates both spatial and temporal information into the segmentation task. A key challenge in training this network is that the available manual annotations are temporally sparse, which forbids end-to-end training. We address this challenge by performing non-rigid label propagation on the annotations and introducing an exponentially weighted loss function for training. Experiments on aortic MR image sequences demonstrate that the proposed method significantly improves both accuracy and temporal smoothness of segmentation, compared to a baseline method that utilises spatial information only. It achieves an average Dice metric of 0.960 for the ascending aorta and 0.953 for the descending aorta.
\end{abstract}

\section{Introduction}
Segmentation is an important task in medical image analysis. It assigns a class label to each pixel/voxel in a medical image so that anatomical structures of interest can be quantified. Recent progress in machine learning has greatly improved the state-of-the-art in medical image segmentation and substantially increased accuracy. However, most of the research so far focuses on static image segmentation, whereas segmentation of temporal image sequences has received less attention. Image sequence segmentation plays an important role in assessing the anatomy and function of moving organs, such as the heart and vessels. In this work, we propose a novel method for medical image sequence segmentation and demonstrate its performance on aortic MR image sequences.

There are two major contributions of this work. First, the proposed method combines a fully convolutional network (FCN) with a recurrent neural network (RNN) for image sequence segmentation. It is able to incorporate both spatial and temporal information into the task. Second, we address the challenge of training the network from temporally sparse annotations. An aortic MR image sequence typically consists of tens or hundreds of time frames. However, manual annotations may only be available for a few time frames. In order to train the proposed network end-to-end from temporally sparse annotations, we perform non-rigid label propagation on the annotations and introduce an exponentially weighted loss function for training.

We evaluated the proposed method on an aortic MR image set from 500 subjects. Experimental results show that the method improves both accuracy and temporal smoothness of segmentation, compared to a state-of-the-art method.

\subsection{Related Works}
\noindent \textbf{FCN and RNN} The FCN was proposed to tackle pixel-wise classification problems, such as image segmentation \cite{Long2015}. Ronnerberger et al. proposed the U-Net, which is a type of FCN that has a symmetric U-shape architecture for feature analysis and synthesis paths \cite{Ronneberger2015}. It has demonstrated remarkable performance in static medical image segmentation. The RNN was designed for handling sequences. The long short-term memory (LSTM) network is a type of RNN that introduces self-loops to enable the gradient flow for long durations \cite{Hochreiter1997}.

In the domain of medical image analysis, the combination of FCN with RNN has been explored recently \cite{ChenJ2016, Poudel2016, YangX2017, KongB2016, XueW2017, HuangW2017}. In some works, RNN was used to model the spatial dependency in static images \cite{ChenJ2016, Poudel2016, YangX2017}, such as the inter-slice dependency in anisotropic images \cite{ChenJ2016, Poudel2016}. In other works, RNN was used to model the temporal dependency in image sequences \cite{KongB2016, XueW2017, HuangW2017}. For example, Kong et al. used RNN to model the temporal dependency in cardiac MR image sequences and to predict the cardiac phase for each time frame \cite{KongB2016}. Xue et al. used RNN to estimate the left ventricular areas and wall thicknesses across a cardiac cycle \cite{XueW2017}. Huang et al. used RNN to estimate the location and orientation of the heart in ultrasound videos \cite{HuangW2017}. These works on medical image sequence analysis \cite{KongB2016, XueW2017, HuangW2017} mainly used RNN for image-level regression. The contribution of our work is that instead of performing regression, we integrate FCN and RNN to perform pixel-wise segmentation for medical image sequences.

\noindent \textbf{Sparse Annotations} Manual annotation of medical images is time-consuming and tedious. It is normally performed by image analysts with clinical knowledge and not easy to outsource. Consequently, we often face small or sparse annotation sets, which is a challenge for training a machine learning algorithm, especially neural networks. To learn from spatially sparse annotations, Cicek et al. proposed to assign a zero weight to unlabelled voxels in the loss function \cite{Cicek2016}. In this work, we focus on learning from temporally sparse annotations and address the challenge by performing non-rigid label propagation and introducing an exponentially weighted loss function.

\noindent \textbf{Aortic Image Segmentation} For aortic image sequence segmentation, a deformable model approach has been proposed \cite{Herment2010}, which requires a region of interest and the centre of aorta to be manually defined in initialisation. This work proposes a fully automated segmentation method.

\begin{figure}[h!]
  \centering
  \includegraphics[width=11.5cm]{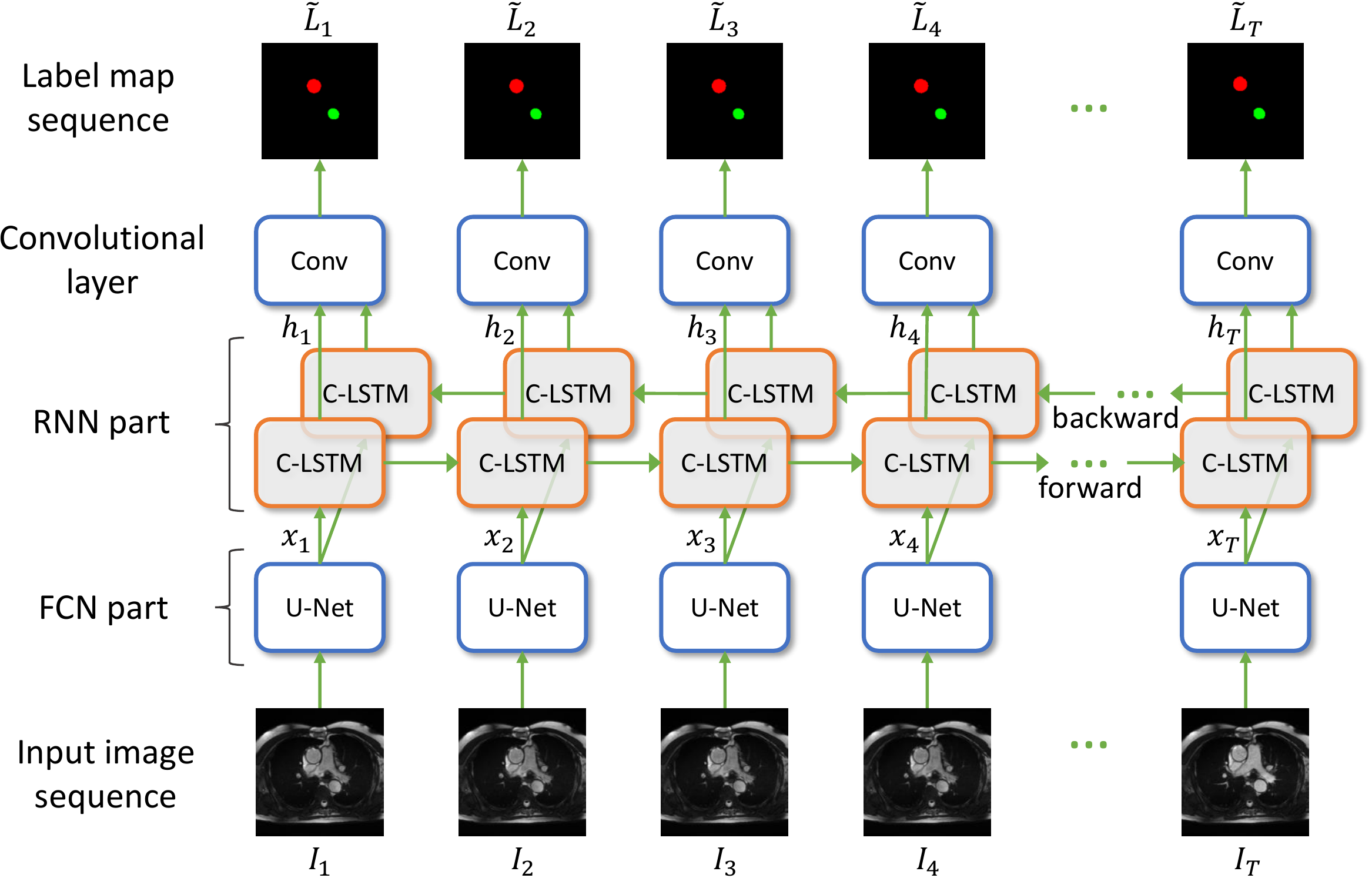}
  \caption{The proposed method analyses spatial features in the input image sequence using U-Net, extracts the second last layer of U-Net as feature maps $x_t$, connects them using convolutional LSTM (C-LSTM) units across the temporal domain and finally predicts the label map sequence. \label{fig:diag}}
\end{figure}

\section{Methods}
\subsection{Network Architecture}
Fig.\ref{fig:diag} shows the diagram of the method. The input is an image sequence $I = \{I_t|t = 1, 2, \dots, T\}$ across time frames $t$ and the output is the predicted label map sequence $\tilde{L} = \{\tilde{L}_t|t = 1, 2, \dots, T\}$. The method consists of two main parts, FCN and RNN. The FCN part analyses spatial features in each input image $I_t$ and extracts a feature map $x_t$. We use the U-Net architecture \cite{Ronneberger2015} for the FCN part, which has demonstrated good performance in extracting features for image segmentation.

The second last layer of the U-Net \cite{Ronneberger2015} is extracted as the feature map $x_t$ and fed into the RNN part. For analysing temporal features, we use the convolutional LSTM (C-LSTM) \cite{Stollenga2015}. Compared to the standard LSTM which analyses one-dimensional signals, C-LSTM is able to analyse multi-dimensional images across the temporal domain. Each C-LSTM unit is formulated as:
\begin{align}
i_t & = \sigma(x_t * W_{xi} + h_{t-1} * W_{hi} + b_i) \nonumber \\
f_t & = \sigma(x_t * W_{xf} + h_{t-1} * W_{hf} + b_f) \nonumber \\
c_t & = c_{t-1} \odot f_t + i_t \odot \tanh(x_t * W_{xc} + h_{t-1} * W_{hc} + b_c) \\
o_t & = \sigma(x_t * W_{xo} + h_{t-1} * W_{ho} + b_o) \nonumber\\
h_t & = o_t \odot tanh(c_t) \nonumber
\end{align}
where $*$ denotes convolution\footnote{The standard LSTM performs multiplication instead of convolution here.}, $\odot$ denotes element-wise multiplication, $\sigma(\cdot)$ denotes the sigmoid function, $i_t$, $f_t$, $c_t$ and $o_t$ are respectively the input gate (i), forget gate (f), memory cell (c) and output gate (o), $W$ and $b$ denote the convolution kernel and bias for each gate, $x_t$ and $h_t$ denote the input feature map and output feature map. The equation shows that output $h_t$ at time point $t$ is determined by both the current input $x_t$ and the previous states $c_{t-1}$ and $h_{t-1}$. In this way, C-LSTM utilises past information during prediction. In the proposed method, we use bi-directional C-LSTM, which consists of a forward stream and a backward stream, as shown in Fig.\ref{fig:diag}, so that the network can utilise both past and future information.

The output of C-LSTM is a pixel-wise feature map $h_t$ at each time point $t$. To predict the probabilistic label map $\tilde{L}_t$, we concatenate the outputs from the forward and backward C-LSTMs and apply a convolution to it, followed by a softmax layer. The loss function at each time point is defined as the cross-entropy between the ground truth label map $L_t$ and the prediction $\tilde{L}_t$.

\begin{figure}[h!]
  \centering
  \subfloat[Label propagation]{
    \includegraphics[width=6cm]{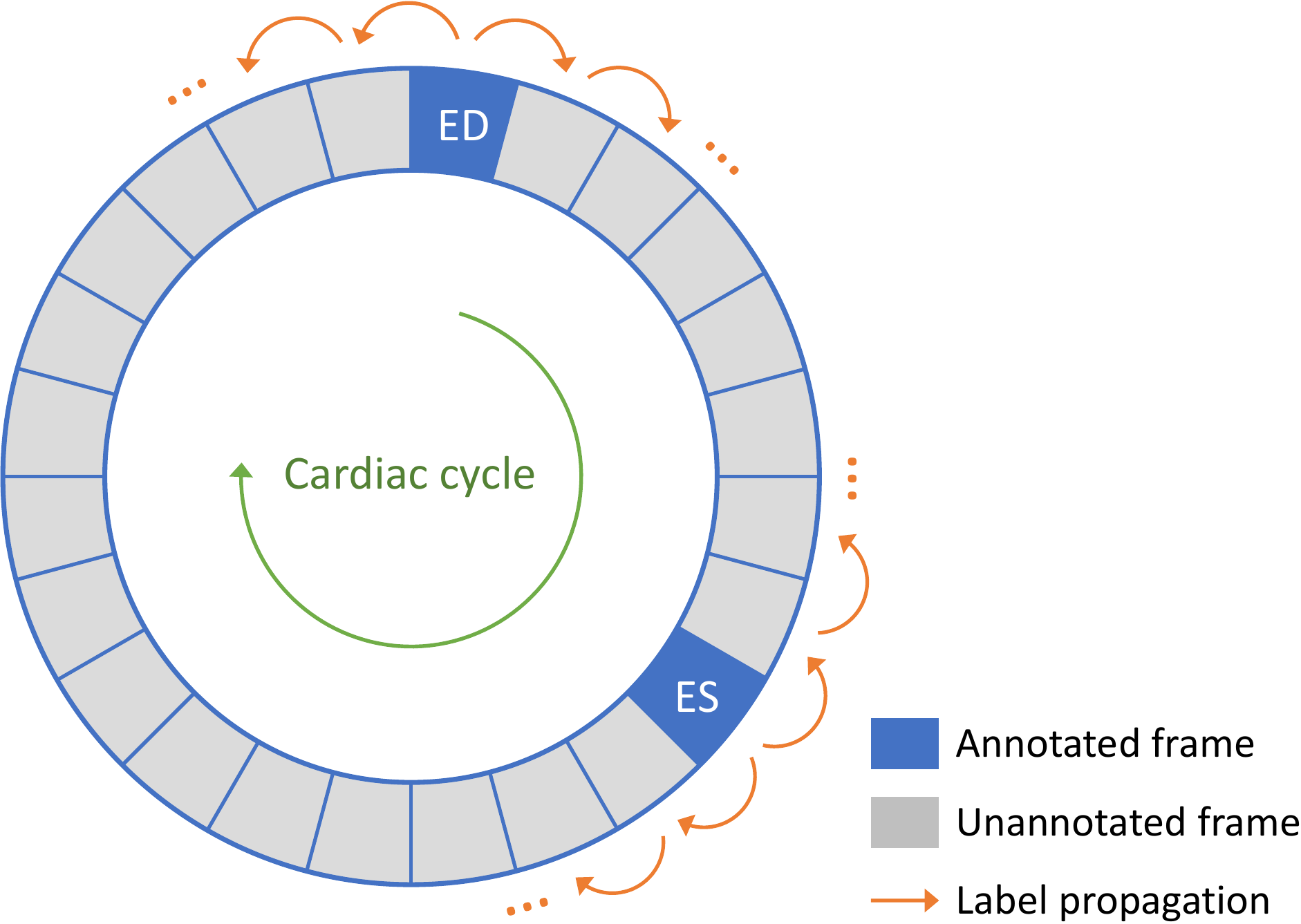}
  }
  \subfloat[Weighting function]{
    \includegraphics[width=5cm]{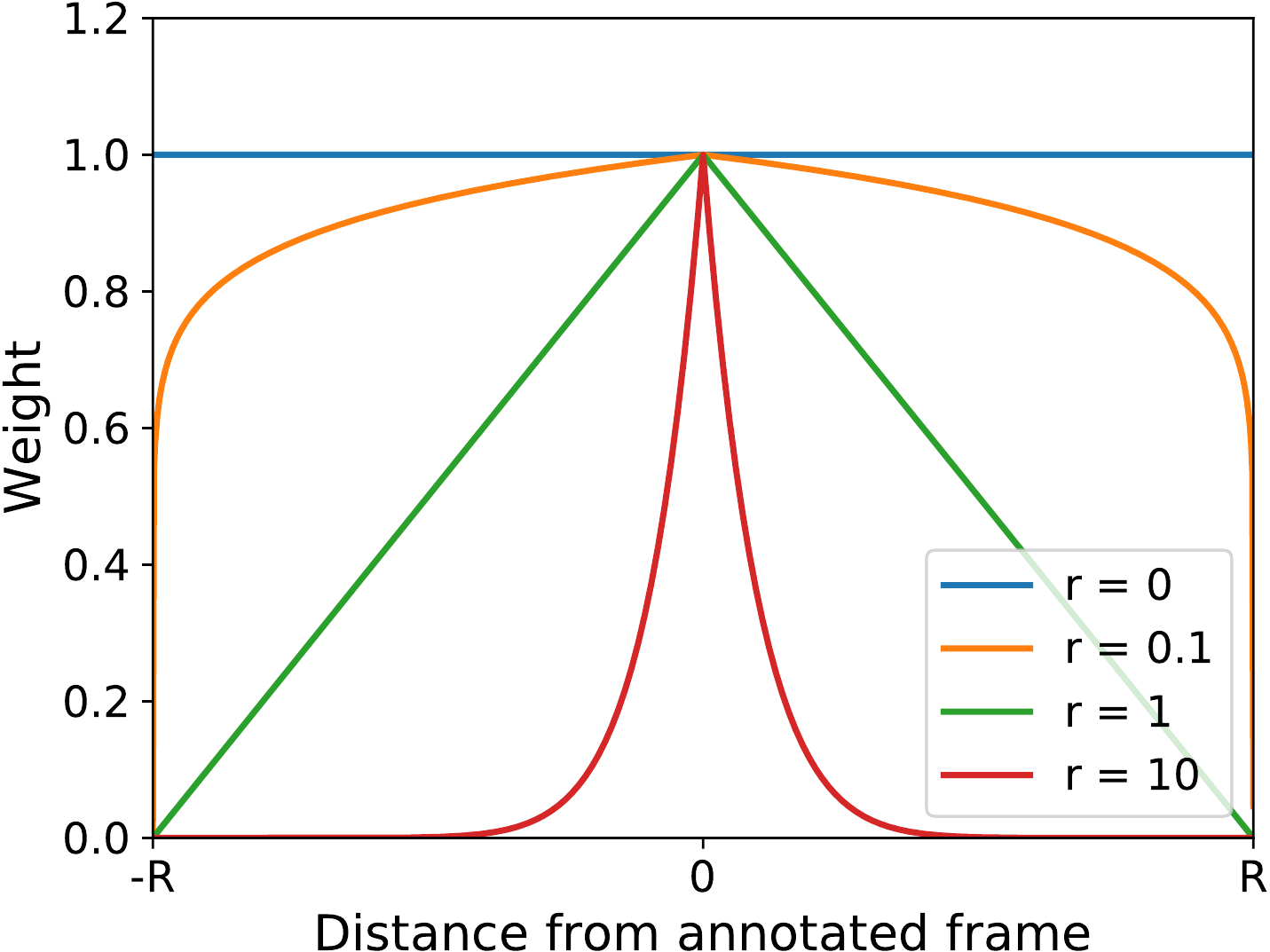}
  }
  \caption{Label propgation and the weighting function for propagated label maps. \label{fig:label_prop}}
\end{figure}

\subsection{Label Propagation and Weighted Loss}
To train the network end-to-end, we require the ground truth label map sequence across the time frames. However, the typical manual annotation is temporally sparse. For example, in our dataset, we only have manual annotations at two time frames, end-diastole (ED) and end-systole (ES). In order to obtain the annotations at other time frames, we perform label propagation. Non-rigid image registration \cite{Rueckert1999} is performed to estimate the motion between each pair of successive time frames. Based on the motion estimate, the label map at each time frame is propagated from either ED or ES annotations, whichever is closer, as shown in Fig.\ref{fig:label_prop}(a).

Registration error may accumulate during label propagation. The further a time frame is from the original annotation, the larger the registration error might be. To account for the potential error in propagated label maps, we introduce a weighted loss function for training,
\begin{equation}
E(\theta) = \sum_{t} w(t - s) \cdot f(L_t, \tilde{L}_t(\theta))
\end{equation}
where $\theta$ denotes the network parameters, $f(\cdot)$ denotes the cross-entropy between the propagated label map $L_t$ and the predicted label map $\tilde{L}_t(\theta)$ by the network, $s$ denotes the nearest annotated time frame to $t$ and $w(\cdot)$ denotes an exponential weighting function depending on the distance between $t$ and $s$,
\begin{equation}
w(t - s) = (1 - \frac{|t - s|}{R})^r
\end{equation}
where $R$ denotes the radius of the time window $T$ for the unfolded RNN and the exponent $r$ is a hyper-parameter which controls the shape of the weighting function. Some typical weighting functions are shown in Fig.\ref{fig:label_prop}(b). If $r = 0$, it treats all the time frames equally. If $r > 0$, it assigns a lower weight to time frames further from the original annotated frame.

\subsection{Evaluation}
We evaluate the method performance in two aspects, segmentation accuracy and temporal smoothness. For segmentation accuracy, we evaluate the Dice overlap metric and the mean contour distance between automated segmentation and manual annotation at ED and ES time frames. We also calculate the aortic area and report the difference between automated measurement and manual measurement. For evaluating temporal smoothness, we plot the curve of the aortic area $A(t)$ against time, as shown in Fig.~\ref{fig:curve}, calculate the curvature of the time-area curve, $\kappa(t) = \frac{|A^{''}(t)|}{(1 + A^{'2}(t))^{1.5}}$, and report the mean curvature across time.

\section{Experiments and Results}
\subsection{Data and Annotations}
We performed experiments on an aortic MR image set of 500 subjects, acquired from the UK Biobank. The typical image size is 240$\times$196 pixel with the spatial resolution of 1.6$\times$1.6 mm$^2$. Each image sequence consists of 100 time frames, covering the cardiac cycle. Two experienced image analysts manually annotated the ascending aorta (AAo) and descending aorta (DAo) at ED and ES time frames. The image set was randomly split into a training set of 400 subjects and a test set of 100 subjects. The performance is reported on the test set.

\subsection{Implementation and Training}
The method was implemented using Python and Tensorflow. The network was trained in two steps. In the first step, the U-Net part was trained for static image segmentation using the Adam optimiser for 20,000 iterations with a batch size of 5 subjects. The initial learning rate was 0.001 and it was divided by 10 after 5,000 iterations. In the second step, the pre-trained U-Net was connected with the RNN and trained together end-to-end using image and propagated label map sequences for 20,000 iterations with the same learning rate settings but a smaller batch size of 1 subject due to GPU memory limit. Data augmentation was performed online, which applied random translation, rotation and scaling to each input image sequence. Training took $\sim$22 hours on a Nvidia Titan Xp GPU. At test time, it took $\sim$10 seconds to segment an aortic MR image sequence.

\subsection{Network Parameters}
There are a few parameters for the RNN, including the length of the time window $T$ after unfolding the RNN and the exponent $r$ for the weighting function. We investigated the impact of these parameters. Table~\ref{tab:par} reports the average Dice metric when the parameters vary. It shows that a combination of time window $T = 9$ and exponent $r = 0.1$ achieves a good performance. When the time window increases to 21, the performance slightly decreases, possibly because the accumulative error of label propagation becomes larger. The exponent $r = 0.1$ outperforms $r = 0$, the latter treating the annotated frames and propagated frames equally, without considering the potential propagation error.

\begin{table}[h!]
  \caption{Mean Dice overlap metrics of the aortas when parameters vary. \label{tab:par}}
  \renewcommand{\arraystretch}{1.1}
  \setlength\tabcolsep{6pt}
  \centering
  \subfloat[Varying $T$ ($r = 0.1$)]{
    \begin{tabular}{c|cc}
    \hline 
    $T$ & AAo & DAo \\ 
    \hline 
    5 & 0.959 & 0.952 \\
    9 & \textbf{0.960} & \textbf{0.953} \\
    13 & 0.959 & 0.950 \\
    17 & 0.959 & 0.952 \\
    21 & 0.958 & 0.951 \\
    \hline 
    \end{tabular} 
  }
  \qquad
  \subfloat[Varying $r$ ($T = 9$)]{
    \begin{tabular}{c|cc}
    \hline 
    $r$ & AAo & DAo \\ 
    \hline 
    0 & 0.955	& 0.949 \\
    0.1 & \textbf{0.960} & \textbf{0.953} \\
    1.0 & 0.959	& 0.951 \\
    10.0 & 0.959 & 0.948 \\
    100.0 & \textbf{0.960} & 0.949 \\
    \hline 
    \end{tabular}
  }
\end{table}

\vspace{-1em}
\begin{table}[h!]
  \caption{Quantitative comparison to U-Net. The columns list the mean Dice metric, contour distance error, aortic area error and time-area curve curvature. \label{tab:comp}}
  \renewcommand{\arraystretch}{1.1}
  \setlength\tabcolsep{6pt}
  \centering
  \begin{tabular}{ccccccccc}
  \hline
  & \multicolumn{2}{c}{Dice metric} & \multicolumn{2}{c}{Dist. error (mm)} & \multicolumn{2}{c}{Area error (mm$^2$)} & \multicolumn{2}{c}{Curvature} \\
  \cmidrule(lr){2-3} \cmidrule(lr){4-5} \cmidrule(lr){6-7} \cmidrule(lr){8-9}
  & AAo & DAo & AAo & DAo & AAo & DAo & AAo & DAo \\ 
  \hline 
  U-Net & 0.953	& 0.944	& 0.80 & 0.69 & 51.68 &	35.96 & 0.47 & 0.38 \\
  Proposed & \textbf{0.960} & \textbf{0.953} & \textbf{0.67} & \textbf{0.59} & \textbf{39.61} & \textbf{27.98} & \textbf{0.41} & \textbf{0.28} \\
  \hline 
  \end{tabular}
\end{table}

\begin{figure}[h!]
  \centering
  \hspace{-0.5cm}
  \includegraphics[width=11.5cm]{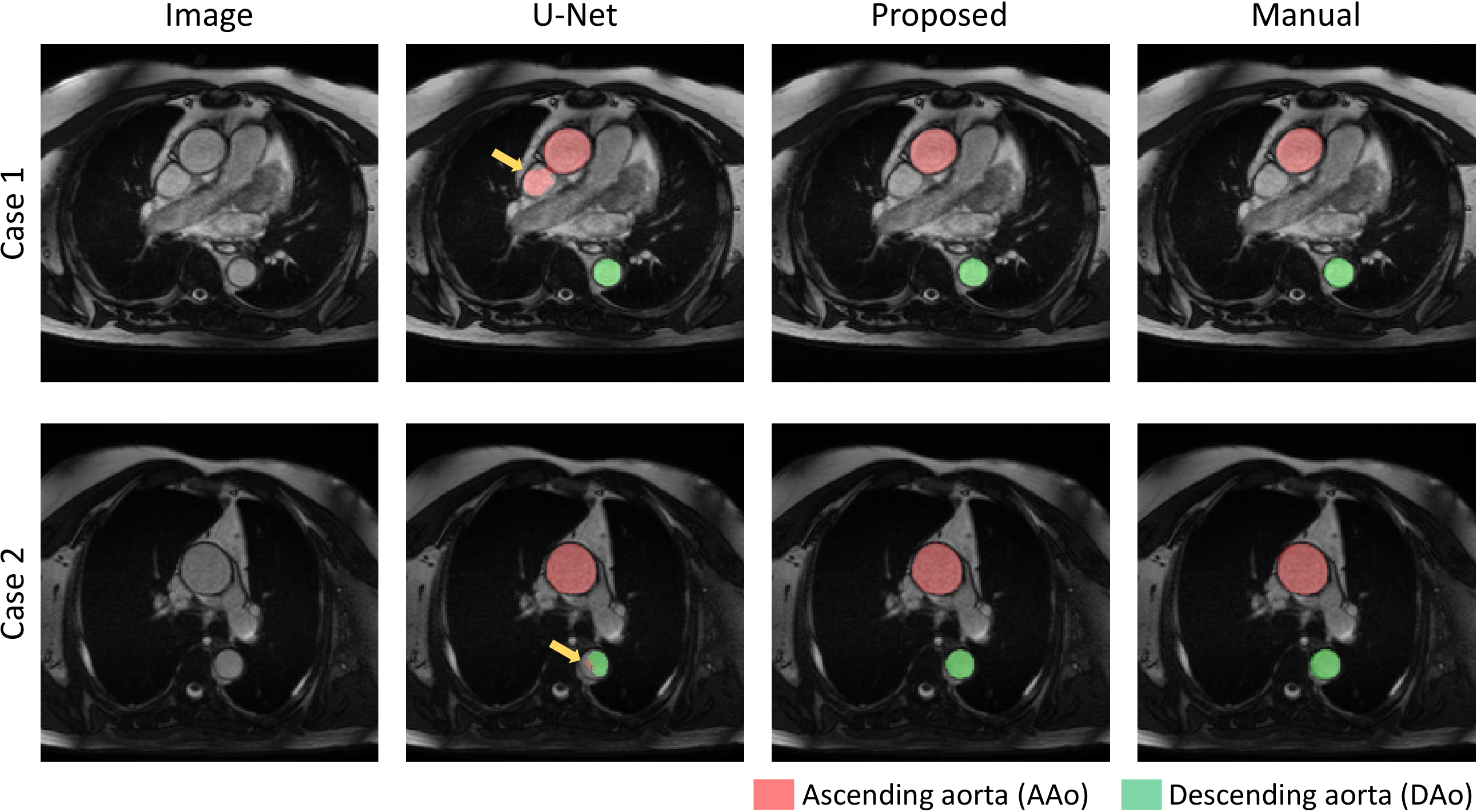}
  \caption{Comparison of the segmentation results for U-Net and the proposed method. The yellow arrows indicate segmentation errors made by U-Net. \label{fig:comp}}
\end{figure}

\begin{figure}[h!]
  \centering
  \vspace{-1.5em}
  \subfloat[Ascending aorta]{
    \includegraphics[width=5.5cm]{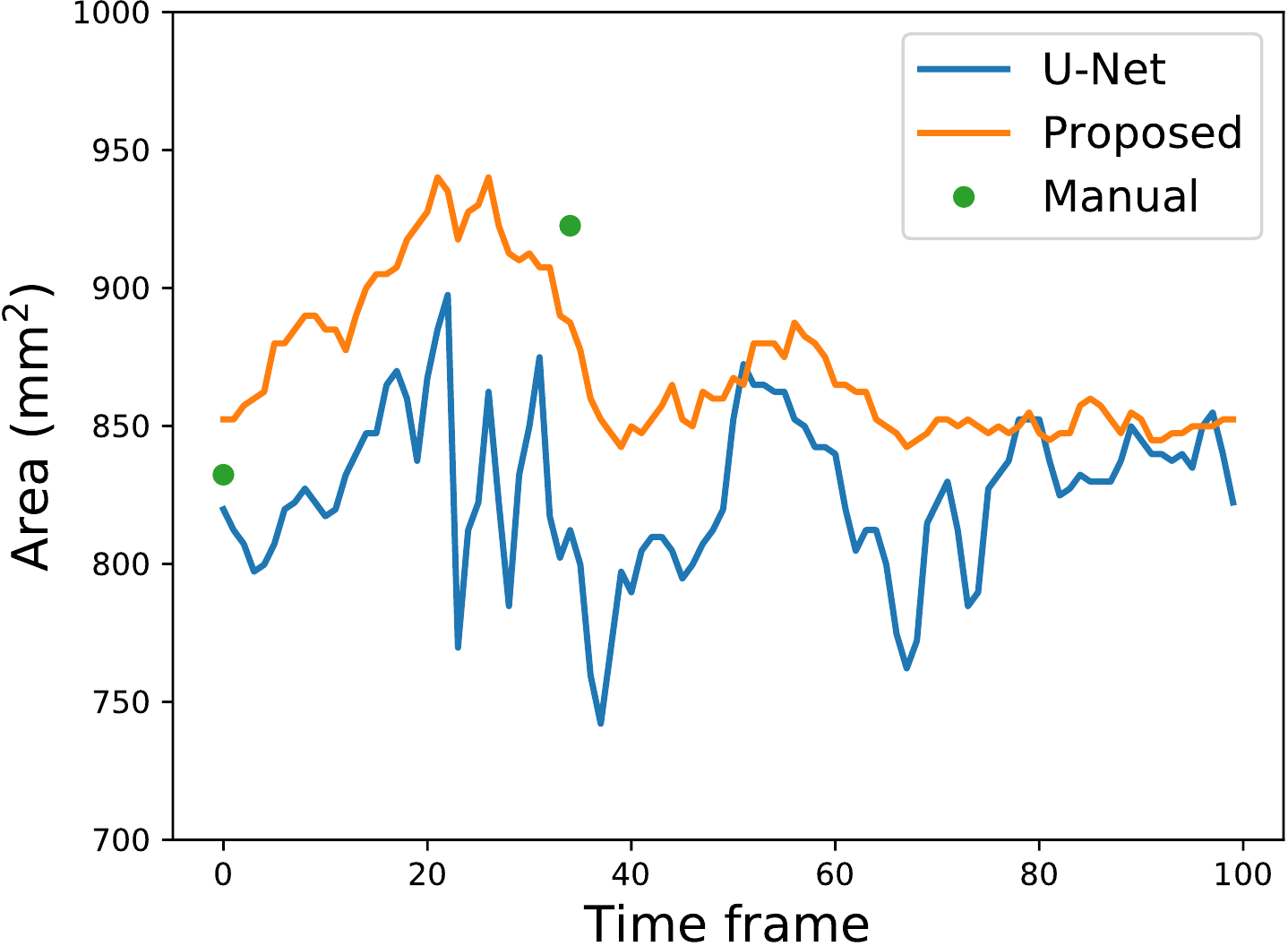}
  }
  \subfloat[Descending aorta]{
    \includegraphics[width=5.5cm]{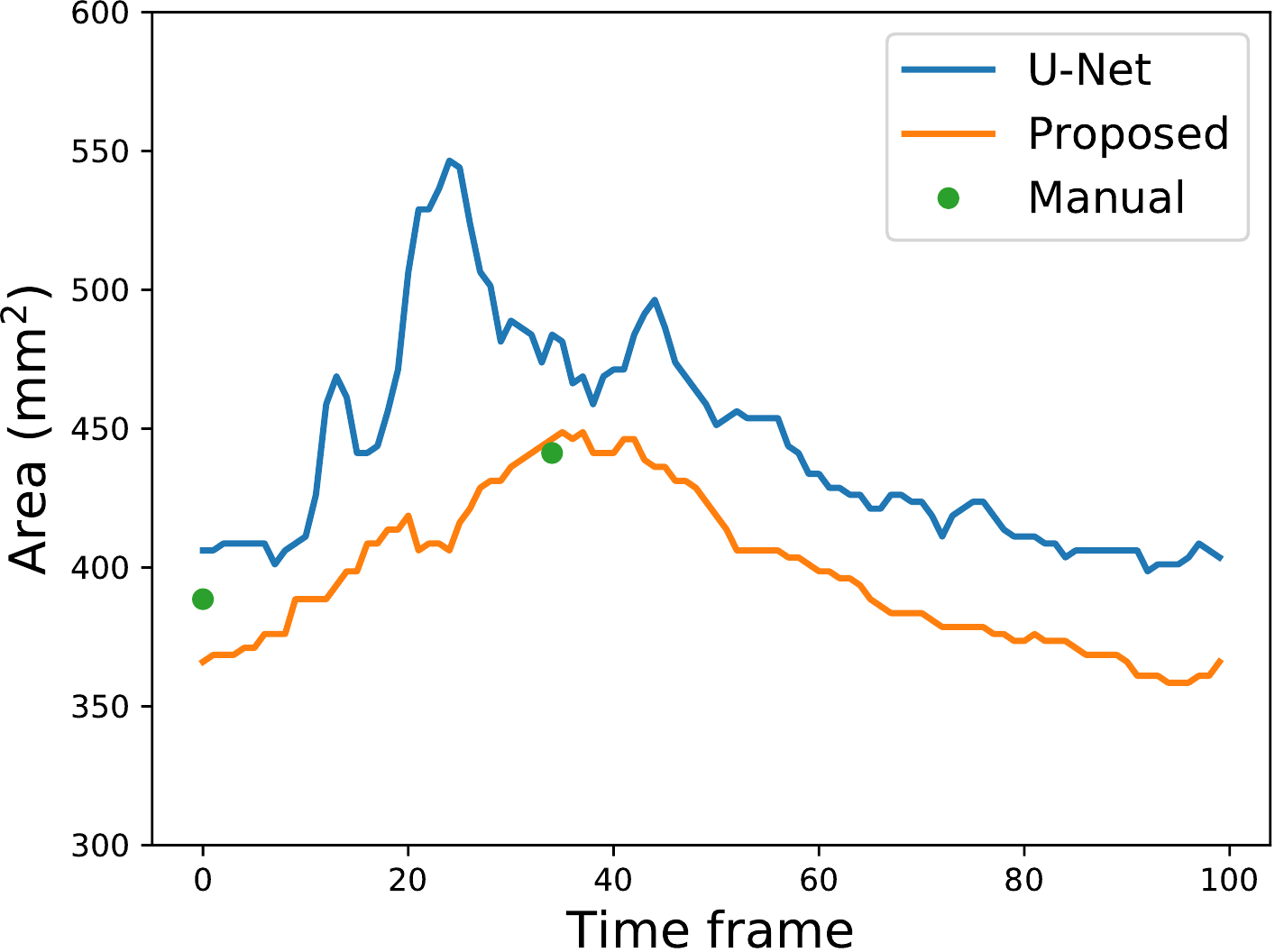}
  }
  \caption{Comparison of aortic time-area curves. The green dots indicate the manual measurements at ED and ES time frames. \label{fig:curve}}
\end{figure}

\subsection{Comparison to Baseline}
We compared the proposed method to the U-Net \cite{Ronneberger2015}, which is a strong baseline method. U-Net was applied to segment each time frame independently. Fig.\ref{fig:comp} compares the segmentation results on two exemplar cases. In Case 1, the U-Net misclassifies a neighbouring vessel as the ascending aorta. In Case 2, the U-Net under-segments the descending aorta. For both cases, the proposed method correctly segments the aortas. Fig.\ref{fig:curve} compares the time-area curves of the two methods on a exemplar subject. It shows that the curve produced by the proposed method is temporally smoother with less abrupt changes. Also, the curve agrees well with the manual measurements at ED and ES. Table~\ref{tab:comp} reports the quantitative evaluation results for segmentation accuracy and temporal smoothness. It shows that the proposed method outperforms the U-Net in segmentation accuracy, achieving a higher Dice metric, a lower contour distance error and a lower aortic area error (all with $p < 0.001$ in paired t-tests). In addition, the proposed method reduces the curvature of the time-area curve ($p < 0.001$), which indicates improved temporal smoothness.

\section{Conclusions}
In this paper, we propose a novel method which combines FCN and RNN for medical image sequence segmentation. To address the challenge of training the network with temporally sparse annotations, we perform non-rigid label propagation and introduce an exponentially weighted loss function for training, which accounts for potential errors in label propagation. We evaluated the method on aortic MR image sequences and demonstrated that by incorporating spatial and temporal information, the proposed method outperforms a state-of-the-art baseline method in both segmentation accuracy and temporal smoothness.

\subsubsection*{Acknowledgements}
This research has been conducted using the UK Biobank Resource under Application Number 18545. This work is supported by the SmartHeart EPSRC Programme Grant (EP/P001009/1). We would like to acknowledge NVIDIA Corporation for donating a Titan Xp for this research. P.M.M. thanks the Edmond J. Safra Foundation, Lily Safra and the UK Dementia Research Institute for their generous support.

\bibliographystyle{unsrt}
\bibliography{refs}

\end{document}